\documentclass{mynote}
\usepackage{url}
\usepackage{fancyvrb}
\usepackage{algorithm}
\usepackage[noend]{algorithmic}
\copyrightyear{2012}
\pubyear{2012}

\begin{document}
\firstpage{1}

\title[A C engine for the MINE algorithms family]{minerva and minepy: a C engine for the MINE suite and its R, Python and MATLAB wrappers}

%\title[A C engine for the MINE algorithms family]{minerva and minepy: MINE analysis of high-throughput multivariate data}

\author[Albanese \textit{et~al}]{
Davide Albanese\,$^{1,+}$,
Michele Filosi\,$^{1,2,+}$, 
Roberto Visintainer\,$^{1,3,+}$,
\\
Samantha Riccadonna\,$^{1}$,
Giuseppe Jurman\,$^{1}$ and 
Cesare Furlanello\,$^{1}$\footnote{to whom correspondence should be addressed
\newline + authors equally contributed to this work }
}

\address{
$^{1}$Fondazione Bruno Kessler, via Sommarive 18, I-38123 Povo (Trento), Italy\\
$^{2}$CIBIO, University of Trento, via delle Regole 101, I-38123 Mattarello (Trento), Italy\\
$^{3}$DISI, University of Trento, Via Sommarive 5, I-38123 Povo (Trento), Italy
}

\history{Received on XXXXX; revised on XXXXX; accepted on XXXXX}

\editor{Associate Editor: XXXXXXX}

\maketitle
\begin{abstract}
\section{Summary:}
We introduce a novel implementation in ANSI C of the MINE family of
algorithms for computing maximal information-based measures of
dependence between two variables in large datasets, with the aim of a
low memory footprint and ease of integration within bioinformatics
pipelines. We provide the libraries minerva (with the R interface) and
minepy for Python, MATLAB, Octave and C++. The C solution reduces the
large memory requirement of the original Java implementation, has good
upscaling properties, and offers a native parallelization for the R
interface. Low memory requirements are demonstrated on the MINE
benchmarks as well as on large ($n$=1340) microarray and Illumina GAII
RNA-seq transcriptomics datasets.

\section{Availability and Implementation:} 
Source code and binaries are freely available for download under GPL3
licence at \url{http://minepy.sourceforge.net} for minepy and through the CRAN repository
\url{http://cran.r-project.org} for the R package minerva.
All software is multiplatform (MS Windows, Linux and OSX).
\section{Contact:} \href{furlan@fbk.eu}{furlan@fbk.eu}
\section{Supplementary information:} 
Supplementary information is available at the website
\url{http://mpba.fbk.eu/minepy-minerva}
\end{abstract}
\section{Introduction}
The Maximal Information-based Nonparametric Exploration (MINE) family
of statistics, including the Maximal Information Coefficient (MIC)
measure, was recently introduced in \citep{reshef11detecting}, aimed
at fast exploration of two-variable relationships in many-dimensional
data sets. MINE consists of the algorithms for computing four measures
of dependence --- MIC, Maximum Asymmetry Score (MAS), Maximum Edge
Value (MEV), Minimum Cell Number (MCN) --- between two variables,
having the generality and equitability property. Generality is the
ability of capturing variable relationships of different nature, while
equitability is the property of penalizing similar levels of noise in
the same way, regardless of the nature of the relation between the
variables. The MINE suite received immediate appraisal as a real
breaktrough in the data mining of complex biological data
\citep{speed11correlation,nature12finding} as well as
criticisms\footnote{See comments and referenced experiments by Simon
  and Tibshirani and by Gorfin et al at
  \url{http://comments.sciencemag.org/content/10.1126/science.1205438}
}. Many groups worldwide have already proposed its use for explorative
data analysis in computational biology, from networks 
dynamics to virus ranking
\citep{weiss12good,das12genome,anderson12ranking,karpinets12analyzing,faust12microbial}.
Together with the algorithm description, the MINE authors provided a
Java implementation (MINE.jar), two wrappers (R and Python), and four
reference datasets \citep{reshef11detecting}. However, applicability
and scalability of MINE.jar on large datasets is currently limited due
to memory requirements and lack of programming interfaces.
%\footnote{\url{http://www.information-management.com/blogs/MIC-MINE-predictive-big-data-Harvard-R-10022590-1.html}}
Further, a native parallelization, currently unavailable, would be of
significant benefit.  These issues are hurdles for a systematic
application of MINE algorithms to high-throughput omics data --- for
example, as a substitute of Pearson correlation in network studies.
Inspired by these considerations, we propose an ANSI C
implementation of the MINE algorithms, and the interfaces for R
(minerva), and for C++, Python and MATLAB/Octave (minepy).
\section{The MINE C engine and its wrappers}
The novel engine (libmine) is written in ANSI C as a clean-room
implementation of the algorithms originally described in
\citep{reshef11detecting}, as the Java source code is not
distributed. Libmine provides three structures describing the data,
the parameter configuration and the maximum normalized mutual
information scores. The core function \texttt{mine\_compute\_score()}
takes a dataset structure and a configuration one as input, returning
a score structure as output, from which four functions compute the
MINE statistics.  The minepy Python module works with Python $\geq
2.6$, with NumPy $\geq 1.3.0$ as the sole requirement: the interface
consists of the class \texttt{minepy.MINE} whose methods match the C
functions.  The R package minerva is built as an R wrapper (R $\geq
2.14$) to the C engine: the main function \texttt{mine} takes the
dataset and the parameter configuration as inputs and returns the four
MINE statistics. Minerva allows native parallelization: based on the R
package \texttt{parallel}, the number of cores can be passed as
parameter to \texttt{mine}, whenever multi-core hardware is
available. The curated version of the CDC15 Spellman yeast dataset
\citep{spellman98comprehensive} used in \citep{reshef11detecting} is
included as example. Documentation (on-line and PDF) for minepy is
available at the minepy website, also as on-line help in R for
minerva.
\paragraph{Performance comparison}
The suite was tested for consistency with MINE.jar v1.0.1 on the
Spellman and microbiome datasets from
\url{http://www.exploredata.net}.
For the Spellman dataset (4381 transcripts and 23 timepoints), MIC
values were computed for all features pairs with MINE.jar and minepy
(both with $\alpha$=0.67). Identical results were found up to five
significance digits: see Fig.~S3 and details in Supplementary
Information (SI). For the microbiome data, for the 77 top ranked
association pairs listed in Tab S13 of \citep{reshef11detecting}, we
obtained 44 identical results and a difference less than 0.01 for
other 29 values (details in SI).
\begin{figure*}[!tb]
\begin{center}
\includegraphics[width=0.80\columnwidth]{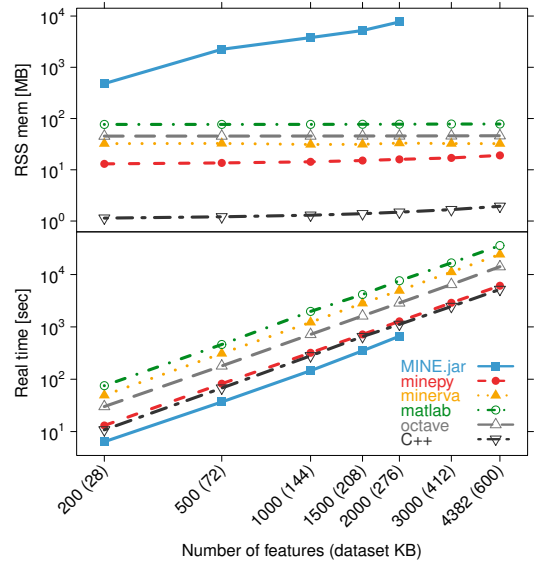}\\
\end{center}
\caption{Comparison of MINE.jar v1.0.1 and the novel interfaces
  (minerva, minepy, Matlab/Octave and C++) computing the four MINE statistics,
  the Pearson correlation coefficient and the non linearity index for
  all pairs of features of the Spellman dataset. For increasing number
  of features: (top) Resident set size (RSS), \textit{i.e.}, the
  non-swapped physical memory (in megabytes); (bottom) elapsed real
  (wall clock) time used by the process, in seconds. MINE.jar can
  complete the task only up to 2000 features, even having reserved 8GB
  RAM to Java. In parentheses, the dataset ASCII file size in kilobytes (KB).}
  \label{fig:spellman-mem}
\end{figure*}
To compare performance for RAM and CPU usage between MINE.jar and the
four wrappers, MINE statistics were computed on all features
pairs of the Spellman dataset, for increasing feature set sizes
(details in SI, Sec 2.2, 2A).
% Analyses were
% repeated 100 times on a single CPU of a 8 cores Intel I7 2.4 GHz 64
% bit workstation equipped with 8Gb RAM and we obtained the same
% numerical results at the third decimal digit.
Minerva and minepy completed all tasks with limited RAM requirements:
about 19 MB were needed for all 4382 variables (600kB dataset size) by
minepy, and 2 MB by the C++ interface.
We were unable to run MINE.jar with more than 2000 variables,
for which Java used 7.5 GB and minepy 16 MB, respectively
(Fig.~\ref{fig:spellman-mem}, Tab.~S2). Minepy computing times are
about twice those of the Java solution (Tab.~S3), but the speedup is
close to 70 for minerva on 100 cores via MPI on a Linux cluster, with
the default $\alpha$=0.6 (Fig.~S6). Finer grids ($\alpha$=0.7) require
much higher computing time as sample size increases (Fig.~S7).

We additionally tested the suite on two recent high-throughput
transcriptomics datasets, of Affymetrix HumanExon 1.0ST human brain
tissues and Illumina Genome Analyzer II sequenced human non-small cell
lung cancer (Tab.~\ref{tab:biodata}). Details on datasets and
experiments are reported in SI.

\begin{table}[!h] 
  \processtable{Performance of minerva and minepy (1-\textit{vs.}-all) on microarray and RNA-seq datasets listed by GEO accession number. $n$: number of samples. $p$: number of features. CPU: Elapsed time used by the process (in seconds). RAM: resident set size (in kilobytes), for minerva (R) and minepy (P).\label{tab:biodata}}
  {\begin{tabular}{crrrrrr} 
      \toprule 
      GEO&  & & \multicolumn{2}{c}{CPU} & \multicolumn{2}{c}{RAM}\\
      Acc. no.& \multicolumn{1}{c}{$n$}& \multicolumn{1}{c}{$p$}&   \multicolumn{1}{c}{R} &  \multicolumn{1}{c}{P} &  \multicolumn{1}{c}{R} & \multicolumn{1}{c}{P} \\ 
      \midrule
      GSE25219$^{1}$ & 1,340 & 17,565& 41,880 & 34,359 & 533,008 & 1,509,692 \\
      GSE34914$^{2}$ & 16   & 20,422   & 42 & 3 & 35,716 & 31,956 \\
      \botrule
    \end{tabular}}{$^1$ \cite{kang11spatio}\; $^2$ \cite{kalari12deep}}
\end{table}

\section*{Acknowledgements}
\paragraph{Funding\textcolon} This work was supported by the EU FP7 Project HiPerDART.
\bibliographystyle{natbib}
\bibliography{albanese12cmine}
\vskip1cm
\appendix{\large{APPENDIX}}
\section{Implementation Details}

The core implementation of minepy and minerva is built from scratch in
ANSI C starting from the pseudocode provided in
\cite{reshef11detecting} Supplementary On-line Material (SOM), as no
original Java source code is available.  The level of detail of the
pseudocode leaves a few ambiguities and in this section we list and
comment the most crucial choices we adopted for the algorithm steps
whenever no explicit description was provided. Obviously, our choices
are not necessarily the same as in the original Java version. The
occurring differences can be ground for small numerical discrepancies
as well as for difference in performance.

\begin{figure*}[!ht]
\begin{tabular}{lr}
\begin{minipage}{0.7\textwidth}
\includegraphics[width=.7\textwidth]{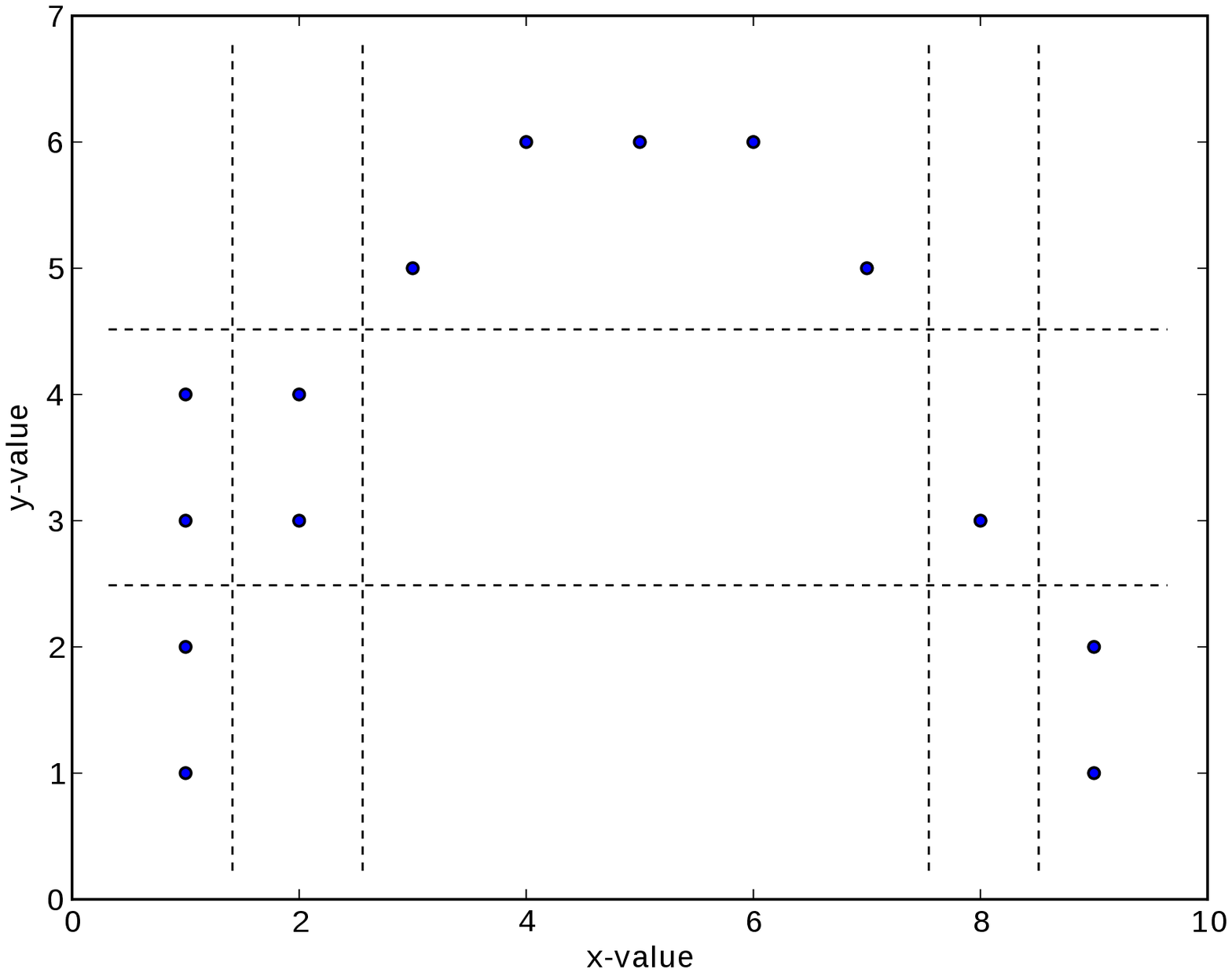} 
\end{minipage}
&
\begin{minipage}{0.3\textwidth}
\begin{tabular}{c|c|c}
\hline
\textbf{D} & \textbf{Q} & \textbf{P} \\
\hline
(1,1) &1&1\\
(1,2) &1&1\\
(1,3) &2&1\\
(1,4) &2&1\\
(2,3) &2&2\\
(2,4) &2&2\\
(3,5) &3&3\\
(4,6) &3&3\\
(5,6) &3&3\\
(6,6) &3&3\\
(7,5) &3&3\\
(8,3) &2&4\\
(9,2) &1&5\\
(9,1) &1&5\\
\hline
\end{tabular}
\end{minipage}
\end{tabular}
\caption{Example of application of the GetClumpsPartition function on
  the set D: first we apply the EquipartitionYAxis(D,y) function with
  $y=3$ and then the GetClumpsPartition(D,Q), obtaining the map
  $P$. Left: the dataset $D$ and the grid cells. Right: mapping of
  points in $D$ for $Q$ ($y$-axis partition) and $P$ ($x$-axis partition).}
 \label{fig:getclumps}
\end{figure*}

\begin{figure*}[!ht]
\begin{center}
\includegraphics[width=0.8\textwidth]{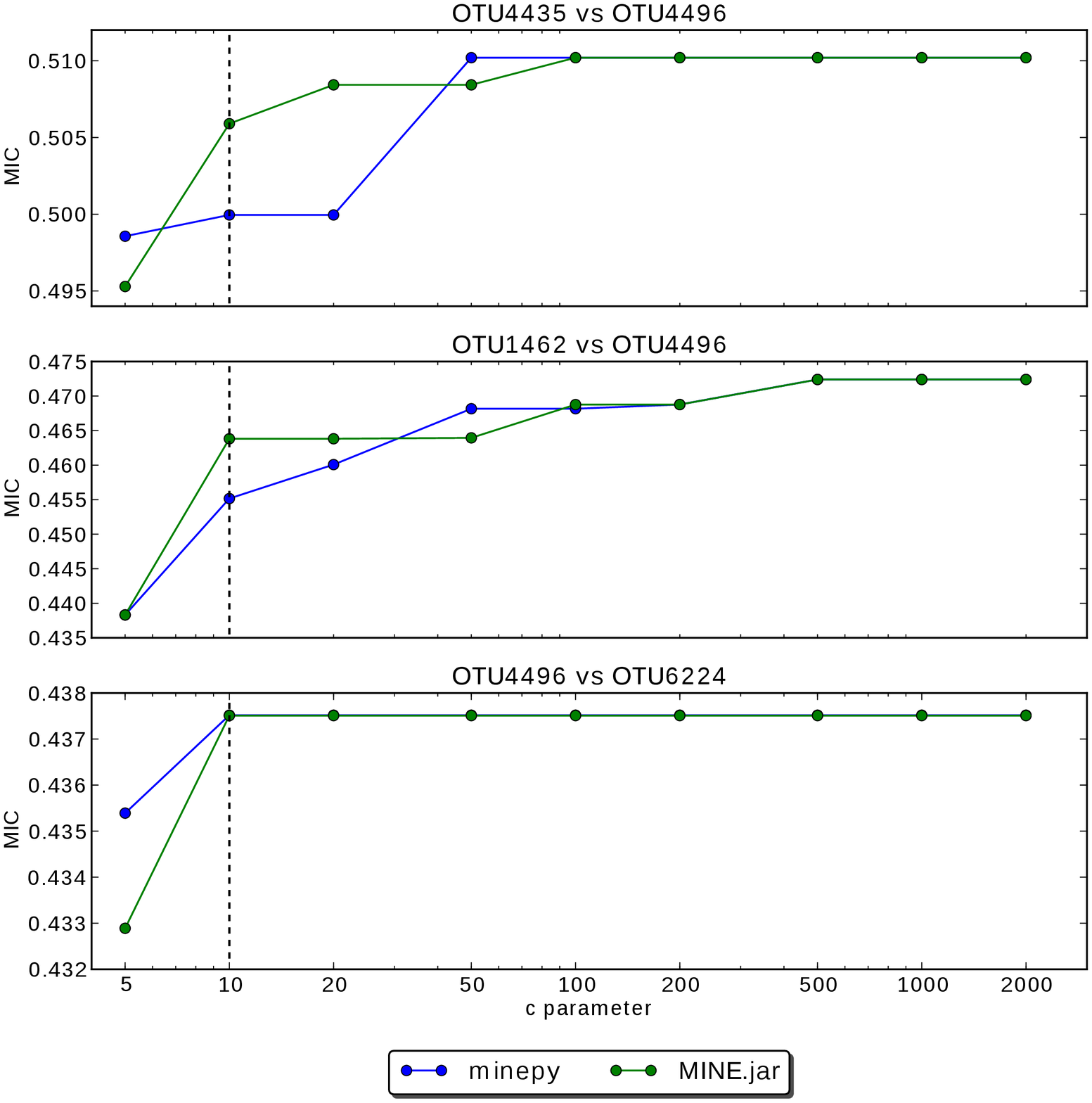} 
\caption{Microbioma dataset. Comparison between MINE.jar and minepy
  for increasing values of the $c$ parameter for the first three
  feature pairs of Tab.~\ref{tab:s13}.  The parameter $c=10$ actually
  used in the experiment is indicated by dashed vertical lines, for
  which MIC differences are found between minepy and MINE.jar (top and middle panels). For larger $c$ values the two
  implementations converge to the same MIC statistic.}
 \label{fig:otucomparison}
\end{center}
\end{figure*}

\begin{enumerate}
\item In SOM, Algorithm 5, the characteristic matrix $M$ is computed
  in the loop starting at line 7 for $xy\leq B$. This is in contrast
  with the definition of the MINE measures (see SOM, Sec. 2) where the
  corresponding bound is $xy<B$ for all the four statistics. We
  adopted the same bound as in the pseudocode, \emph{i.e.} $xy\leq B$.
\item The MINE statistic  MCN is defined as follows in SOM, Sec. 2:
\begin{displaymath}
\textrm{MCN}(D,\epsilon) = \min_{xy<B} \{ \log(xy)\colon M(D)_{x,y} \geq (1-\epsilon)\textrm{MIC}(D)\}
\end{displaymath}
As for MINE.jar (inferred from Table S1), we set $\epsilon=0$ and
$\log$ to be in base 2. Finally, as specified in Point 1 above, we use
the bound $xy\leq B$ as in the SOM pseudocode rather than the $xy<B$
as in the definition.  This led to implement the formula:
\begin{displaymath}
\textrm{MCN}(D,0) = \min_{xy\leq B} \{ \log_2(xy)\colon M(D)_{x,y} = \textrm{MIC}(D)\}\, 
\end{displaymath}
being $\textrm{MIC}(D)$ the maximum value of the matrix $M(D)$.
\item In EquipartitionYAxis() (SOM, Algorithm 3, lines 4 and 10), two
  ratios are assigned to the variable desiredRowSize, namely
  $\frac{n}{y}$ and $\frac{(n-i+1)}{(y-\textrm{currRow}+1)}$. We
  choose to consider the ratios as real numbers; a possible
  alternative is to cast desiredRowSize to an integer. The two
  alternatives can give rise to different $Q$ maps, and thus to
  slightly different numerical values of the MINE statistics.
\item In some cases, the function EquipartitionYAxis() can return a map
  $Q$ whose number of clumps $\hat{y}$ is smaller than $y$,
  \textit{e.g.} when in $D$ there are enough points whose second
  coordinates coincide. This can lead to underestimate the normalized
  mutual information matrix $M_{x,y}$ (SOM, Algorithm 5, line 9),
  where  $M_{x,y}$ is obtained
  by dividing the mutual information $I_{x,y}$ for $\min\{\log x,\log
  y\}$. To prevent this issue, we normalize instead by the factor
  $\min\{\log x,\log \hat{y}\}$.

\item The function GetClumpsPartition($D,Q$) is discussed
  (\cite{reshef11detecting}, SOM page 12), but its pseudocode is not
  explicitely available. Our implementation is defined here in
  Alg.~\ref{alg:GetClumpsPartition}. The function returns the map $P$
  defining the clumps for the set $D$, with the constraint of keeping
  in the same clump points with the same $x$-value. An example of $P$
  partition produced by GetClumpsPartition on a simple set $D$ is
  given in Fig.~\ref{fig:getclumps}.

\item We also explicitly provide the pseudocode for the
  GetSuperclumpsPartition() function (discussed in
  \cite{reshef11detecting}, SOM page 13) in
  Alg.~\ref{alg:GetSuperclumpsPartition}. This function limits the
  number of clumps when their number $k$ is larger than a given bound
  $\hat{k}$.  The function calls the GetClumpsPartition and, for
  $k>\hat{k}$ it builds an auxiliary set $D_{\tilde{P}}$ as an input
  for the EquipartitionYAxis function discussed above (Points 3-4).

\item We observed that the GetSuperclumpsPartition() implemented in
  MINE.jar may fail to respect the $\hat{k}$ constraints on the
  maximum number of clumps and a map $P$ with $\hat{k}+1$ superclumps
  is actually returned. As an example, the MINE.jar applied in debug
  mode (d=4 option) with the same parameters ($\alpha=0.551$, $c=10$)
  used in \cite{reshef11detecting} to the pair of variables
  (OTU4435,OTU4496) of the Microbioma dataset, returns $cx+1$ clumps,
  instead of stopping at the bound $\hat{k}=cx$ for
  $x=12,7,6,5,4\ldots$

\item The possibly different implementations of the
  GetSuperclumpsPartition() function described in Points 6-7 can lead
  to minor numerical differences in the MIC statistics, as documented
  in Section 2.1. To confirm this effect, we verified that by reducing
  the number of calls to the GetSuperclumpsPartition() algorithm, we
  can also decrease the difference between MIC computed by minepy and
  by MINE.jar, and they asymptotically converge to the same value.
  This effect is displayed in Fig. \ref{fig:otucomparison}, where it
  is obtained by increasing the value of $c$ in the MIC computation.

\item In our implementation, we use double-precision floating-point
  numbers (\texttt{double} in C) in the computation of entropy and
  mutual information values. The internal implementation of the same
  quantities in MINE.jar is unknown. 

\item In order to speed up the computation of the MINE statistics, we
  introduced two improvements (with respect to the pseudo-code), in
  OptimizeXAxis(), defined in Algorithm 2 in
  \cite{reshef11detecting} SOM):
  \begin{enumerate}
  \item Given a $(P,Q)$ grid, we precalculate the matrix of number of
    samples in each cell of the grid, to speed up the computation of
    entropy values  $H(Q)$, $H(\langle c_0,c_s,c_t\rangle)$, $H(\langle
    c_0,c_s,c_t \rangle, Q)$ and $H(\langle c_s,c_t \rangle, Q)$
  \item We precalculate the entropy matrix $H(\langle c_s,c_t \rangle,
    Q), \forall s, t$ to speed up the computation of $F(s,t,l)$ (see
    Algorithm 2, lines 10--17 in \cite{reshef11detecting} SOM).
  \end{enumerate}
  These improvements do not affect the final results of mutual
  information matrix and of MINE statistics.
\end{enumerate}

\algsetup{indent=1.5em}
\begin{algorithm}
 \caption{GetClumpsPartition($D,Q$)}
 \label{alg:GetClumpsPartition}
 \begin{algorithmic}[1]
	\REQUIRE $D=\{(a_i,b_i), i=1,\dots,n\}$ is a set of $n$ ordered pairs sorted in increasing order by their first component $a_i$
	\REQUIRE $Q$ is the map of row assignments returned by EquipartitionYAxis
	\ENSURE Returns a map $P:D \rightarrow \{1,\dots,k\}$ providing the column assignment of the point $(a,b)$
	%\NoNumber{}
	\STATE $\tilde{Q} \leftarrow Q$
	\STATE $i \leftarrow 1$
	\STATE $c \leftarrow -1$
	%\NoNumber{}
	\REPEAT
		\STATE $s \leftarrow 0$
		\STATE $\textit{flag} \leftarrow \FALSE$
		\FOR{$j=i+1$ \TO $n$}
			\IF{$a_i = a_j$} 
				\STATE $s \leftarrow s+1$
				\IF{$\tilde{Q}((a_i,b_i)) \neq \tilde{Q}((a_j,b_j))$}
					\STATE $\textit{flag} \leftarrow \TRUE$
				\ENDIF
			\ENDIF
		\ENDFOR
		\IF{$s \neq 0$ \AND {\textit{flag}}}
			\FOR{$j=0$ \TO $s$}
				\STATE $\tilde{Q}((a_{i+j},b_{i+j})) \leftarrow c$
				\STATE $c \leftarrow c-1$
			\ENDFOR
		\ENDIF
		\STATE $i \leftarrow i+s+1$
	\UNTIL{$i > n$}
	%\NoNumber{ }
	\STATE $i \leftarrow 1$
	\STATE $P((a_1,b_1)) \leftarrow i$
	\FOR{$j=2$ \TO $n$}
		\IF{$\tilde{Q}((a_j,b_j)) \neq \tilde{Q}((a_{j-1},b_{j-1}))$}
			\STATE $i \leftarrow i+1$
		\ENDIF
		\STATE $P((a_j,b_j)) \leftarrow i$
	\ENDFOR
	\RETURN $P$
\end{algorithmic}
\end{algorithm}

\algsetup{indent=1.5em}
\begin{algorithm}
 \caption{GetSuperclumpsPartition($D,Q,\hat{k}$)}
 \label{alg:GetSuperclumpsPartition}
 \begin{algorithmic}[1]
	\REQUIRE $D=\{(a_i,b_i), i=1,\dots,n\}$ is a set of $n$ ordered pairs sorted in increasing order by their first component $a_i$
	\REQUIRE $Q$ is the map of row assignments returned by EquipartitionYAxis
	\REQUIRE $\hat{k}$ is the maximum number of clumps
	\ENSURE Returns a map $P:D \rightarrow \{1,\dots,k\}$ providing the column assignment of the point $(a,b)$
	\STATE $\tilde{P} \leftarrow$ GetClumpsPartition($D,Q$)
	\STATE $k \leftarrow$ number of clumps of $\tilde{P}$
	\IF{$k > \hat{k}$}
		\STATE $D_{\tilde{P}} \leftarrow \{(0,\tilde{P}((a_i,b_i))): (a_i,b_i) \in D \}$
%         	\FOR{$i=1$ \TO $n$}
%			\STATE $D_P \leftarrow (-, P((a_i, b_i)))$
%		\ENDFOR
		\STATE $ \hat{P} \leftarrow $ EquipartitionYAxis($D_{\tilde{P}}, \hat{k}$)
	         \STATE $P((a_i,b_i)) \leftarrow \hat{P}((0,\tilde{P}((a_i,b_i))))$ for every $(a_i,b_i)$
		\RETURN $P$
	\ELSE
		\RETURN $\tilde{P}$
	\ENDIF
\end{algorithmic}
\end{algorithm}

% \begin{algorithm}
%  \caption{GetSuperclumpsPartition($D,Q,\hat{k}$)}
%  \label{alg:GetSuperclumpsPartition}
%  \begin{algorithmic}[1]
% 	\REQUIRE $D=\{(a_i,b_i), i=1,\dots,n\}$ is a set of $n$ ordered pairs sorted in increasing order by their first component $a_i$
% 	\REQUIRE $Q$ is the map of row assignments returned by EquipartitionYAxis
% 	\REQUIRE $\hat{k}$ is the maximum number of clumps
% 	\ENSURE Returns a map $P:D \rightarrow \{1,\dots,k\}$ providing the column assignment of the point $(a,b)$
% 	\STATE $\tilde{P} \leftarrow$ GetClumpsPartition($D,Q$)
% 	\STATE $k \leftarrow$ number of clumps of $\tilde{P}$
% 	\IF{$k > \hat{k}$}
% 		\STATE $D_{\tilde{P}} \leftarrow \{(0,\tilde{P}((a_i,b_i))): (a_i,b_i) \in D \}$
% %         	\FOR{$i=1$ \TO $n$}
% %			\STATE $D_P \leftarrow (-, P((a_i, b_i)))$
% %		\ENDFOR
% 		\STATE $P \leftarrow $ EquipartitionYAxis($D_{\tilde{P}}, \hat{k}$)
% 		\RETURN $P$
% 	\ELSE
% 		\RETURN $\tilde{P}$
% 	\ENDIF
% \end{algorithmic}
% \end{algorithm}

%\begin{algorithm}
% \caption{EquipartitionYAxis}
% \label{alg:EquipartitionYAxis}
%\end{algorithm}
\section{Comparison with MINE.jar}

\subsection{Consistency}

\begin{figure*}[!ht]
  \centerline{\includegraphics[width=0.75\columnwidth]{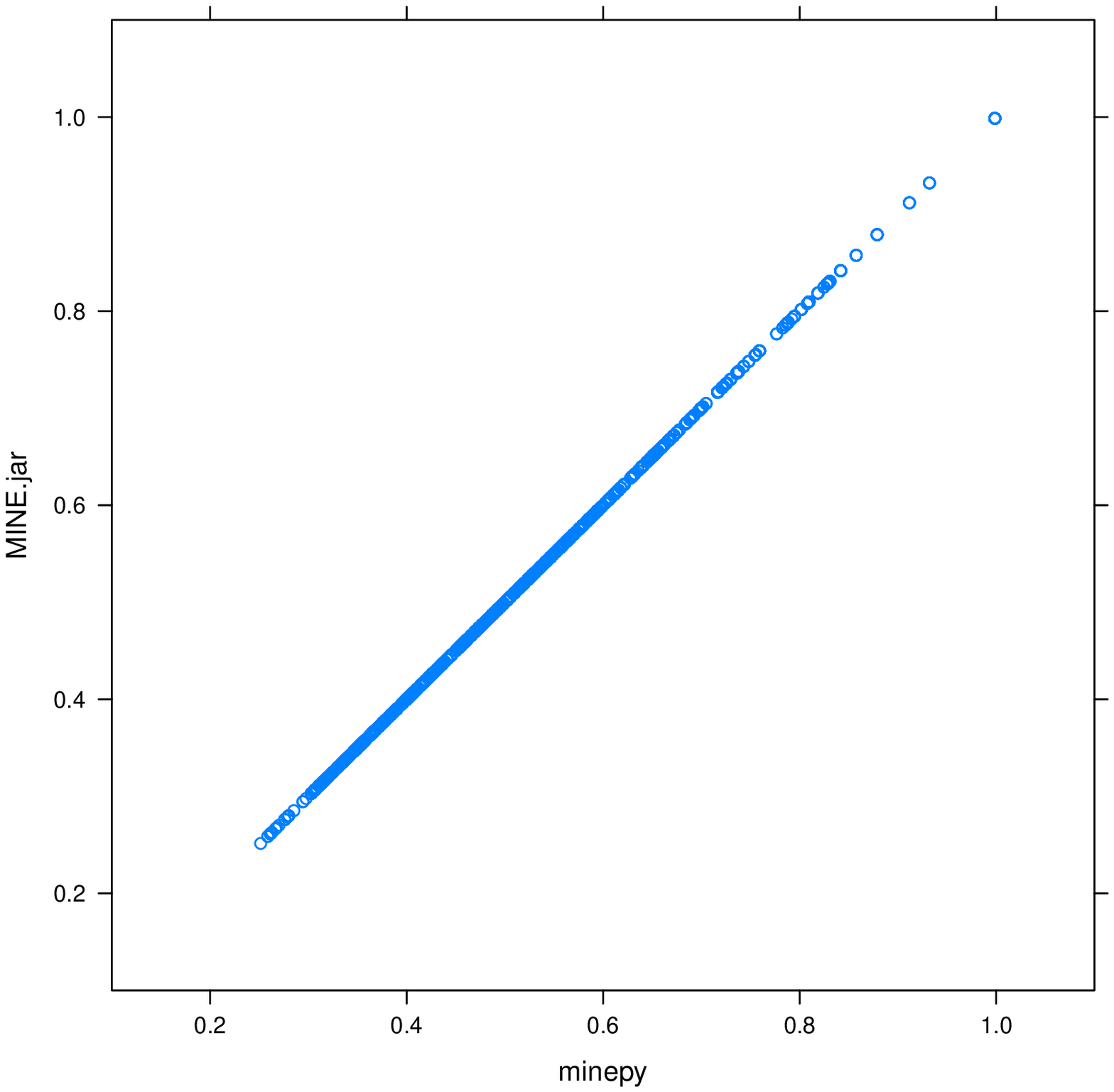}}
  \caption{A comparison of MIC values for minepy (x axis) and the
    original MINE.jar (y axis) on the Spellman dataset (time
    \textit{vs.}  all the other 4381 variables). Parameters are
    $\alpha$=0.67 and $c$$=$15 for both implementations.}
  \label{fig:spellman-mic}
\end{figure*}

\noindent
\paragraph{1A} Minepy was compared with the original MINE.jar
implementation on a ``one vs all'' MINE association study on the
Spellman dataset\footnote{See Subsection \ref{ssec:spellman} for the
  script code of the MINE analysis on this dataset.}. In this task,
MINE statistics were computed for variable \#1 (time) \textit{vs.} all
the other 4381 variables, with grid parameter $\alpha$$=$0.67 and
$c$$=$15, as in \cite {reshef11detecting}. MIC values for the two
implementations are compared in Fig.~\ref{fig:spellman-mic}: all
values coincide up to five digit precision, \emph{i.e.} the precision
of MINE.jar output. Specifications of hardware and software
configurations for this and the other experiments described in this
document are summarized in Sec.~\ref{sec:system}.

\noindent
\paragraph{1B} The MINE analysis was run for all feature pairs with
minepy on the microbiome dataset (6696 features) with $\alpha=0.551$
and $c=10$, as chosen in \cite{reshef11detecting}.  MIC
values\footnote{ A complete table of minepy computed MIC values is
  downloadable at\\
  \url{http://sourceforge.net/projects/minepy/files/data/Microbiome_MIC_minepy_v0.3.4.csv.zip}.}
were compared with the 77 top ranked associations from Tab S13,
Supp. Mat. in \cite{reshef11detecting}. MIC values are identical in 44
cases, for 73 the difference is less than 0.01. The median of all
differences is 0, the 3rd quartile is 0.003, and the largest observed
difference is 0.014; all comparisons are listed in Tab.~\ref{tab:s13}.

\begin{table*}[!ht]
\footnotesize
\begin{center}
\begin{tabular}{llrr|llrr}
\multicolumn{1}{c}{\textbf{OTUX}} & \multicolumn{1}{c}{\textbf{OTUY}}& \multicolumn{1}{c}{\textbf{J}} 
& \multicolumn{1}{c|}{\textbf{P}} &
\multicolumn{1}{c}{\textbf{OTUX}} & \multicolumn{1}{c}{\textbf{OTUY}}& 
\multicolumn{1}{c}{\textbf{J}} & \multicolumn{1}{c}{\textbf{P}} \\
\hline
  OTU4435 & OTU4496 & 0.506 & 0.500 & OTU453 & OTU4496 & 0.282 & 0.284 \\ 
  OTU1462 & OTU4496 & 0.464 & 0.455 & OTU1347 & OTU5937 & 0.300 & 0.290 \\ 
  OTU4496 & OTU6224 & 0.438 & 0.438 & OTU2728 & OTU5937 & 0.276 & 0.276 \\ 
  OTU155 & OTU4496 & 0.425 & 0.425 & OTU2970 & OTU6256 & 0.289 & 0.289 \\ 
  OTU4496 & OTU5417 & 0.414 & 0.414 & OTU1629 & OTU3991 & 0.292 & 0.292 \\ 
  OTU675 & OTU5937 & 0.414 & 0.412 & OTU4865 & OTU5937 & 0.321 & 0.321 \\ 
  OTU2728 & OTU4496 & 0.408 & 0.408 & OTU3991 & OTU4273 & 0.266 & 0.266 \\ 
  OTU5417 & OTU5937 & 0.408 & 0.410 & OTU2350 & OTU6256 & 0.272 & 0.272 \\ 
  OTU4273 & OTU4496 & 0.374 & 0.372 & OTU2941 & OTU6256 & 0.280 & 0.280 \\ 
  OTU675 & OTU6256 & 0.362 & 0.357 & OTU5420 & OTU5937 & 0.432 & 0.431 \\ 
  OTU1629 & OTU6256 & 0.374 & 0.374 & OTU4496 & OTU4501 & 0.259 & 0.259 \\ 
  OTU2970 & OTU4496 & 0.358 & 0.357 & OTU1285 & OTU4496 & 0.256 & 0.256 \\ 
  OTU4273 & OTU5937 & 0.366 & 0.366 & OTU5407 & OTU5420 & 0.272 & 0.275 \\ 
  OTU710 & OTU4496 & 0.354 & 0.354 & OTU3991 & OTU4435 & 0.251 & 0.251 \\ 
  OTU4257 & OTU6256 & 0.346 & 0.355 & OTU5826 & OTU6256 & 0.251 & 0.252 \\ 
  OTU4257 & OTU4496 & 0.339 & 0.325 & OTU1285 & OTU5937 & 0.296 & 0.286 \\ 
  OTU1193 & OTU4496 & 0.336 & 0.342 & OTU4496 & OTU5826 & 0.255 & 0.255 \\ 
  OTU2642 & OTU2728 & 0.385 & 0.383 & OTU4865 & OTU6256 & 0.249 & 0.249 \\ 
  OTU1373 & OTU4496 & 0.322 & 0.322 & OTU6256 & OTU6484 & 0.247 & 0.247 \\ 
  OTU4273 & OTU6256 & 0.329 & 0.335 & OTU4496 & OTU6484 & 0.247 & 0.247 \\ 
  OTU1462 & OTU3991 & 0.326 & 0.326 & OTU1629 & OTU5937 & 0.266 & 0.267 \\ 
  OTU2941 & OTU4496 & 0.332 & 0.332 & OTU4865 & OTU5407 & 0.245 & 0.245 \\ 
  OTU2728 & OTU5490 & 0.353 & 0.353 & OTU4496 & OTU4865 & 0.252 & 0.252 \\ 
  OTU4496 & OTU5420 & 0.312 & 0.318 & OTU2399 & OTU2728 & 0.244 & 0.244 \\ 
  OTU1629 & OTU4496 & 0.305 & 0.305 & OTU2399 & OTU5420 & 0.250 & 0.250 \\ 
  OTU5937 & OTU6224 & 0.301 & 0.293 & OTU2516 & OTU4496 & 0.253 & 0.253 \\ 
  OTU2350 & OTU4496 & 0.302 & 0.302 & OTU5117 & OTU6256 & 0.243 & 0.243 \\ 
  OTU1347 & OTU6256 & 0.329 & 0.320 & OTU5826 & OTU5937 & 0.264 & 0.264 \\ 
  OTU675 & OTU4496 & 0.296 & 0.294 & OTU4435 & OTU5937 & 0.244 & 0.245 \\ 
  OTU4496 & OTU5117 & 0.291 & 0.291 & OTU2728 & OTU5407 & 0.239 & 0.242 \\ 
  OTU1285 & OTU6256 & 0.287 & 0.287 & OTU2036 & OTU6256 & 0.236 & 0.236 \\ 
  OTU1347 & OTU4496 & 0.291 & 0.286 & OTU774 & OTU1347 & 0.236 & 0.236 \\ 
  OTU4496 & OTU5370 & 0.291 & 0.290 & OTU2970 & OTU3991 & 0.236 & 0.232 \\ 
  OTU3994 & OTU5937 & 0.285 & 0.277 & OTU3991 & OTU5370 & 0.235 & 0.235 \\ 
  OTU3994 & OTU4496 & 0.294 & 0.294 & OTU155 & OTU6256 & 0.251 & 0.251 \\ 
  OTU4257 & OTU5937 & 0.379 & 0.369 & OTU4501 & OTU6256 & 0.232 & 0.224 \\ 
  OTU710 & OTU5937 & 0.296 & 0.290 & OTU453 & OTU3991 & 0.241 & 0.235 \\ 
  OTU1373 & OTU5937 & 0.281 & 0.281 & OTU1548 & OTU4496 & 0.234 & 0.234 \\ 
  OTU1548 & OTU6256 & 0.277 & 0.277 &  &  &  &  \\ 
\end{tabular}
\normalsize
\vskip 0.25in
\caption{Side by side comparison (J: MINE.jar; P: minepy) for the 77 top
  ranked MIC (OTUX,OTUY) pairs from Tab. S13 of \cite{reshef11detecting}
  SOM.}
\label{tab:s13}
\end{center}
\end{table*}

\subsection{Performance}

\paragraph{2A.} We compared the computing performance of the C MINE
wrappers (minepy, minerva, the MATLAB/Octave and the C++ interfaces) with
MINE.jar on the CDC15 Spellman dataset. A MINE analysis (four MINE
statistics, Pearson and non-linearity) was run on all feature pairs
for all the 23 time points and increasing feature set sizes. Results
are for memory and elapsed computing times are summarized in Fig.~1
(main paper) and here reported in Tab.~\ref{tab:mem} and
Tab.~\ref{tab:real.times} respectively.
\par
In terms of computing time, MINE.jar has a good performance (See
Fig. 1, bottom panel), but it requires a much larger amount of
memory. With the default parameters, which correspond to reserving 3GB
to the process, we were unable to complete tasks for 1,500 or more
features with MINE.jar (OutOfMemoryError exception). On the 12 GB RAM
system used in this experiment, we then reserved 8 GB by using the -Xmx
option (suggested on MINE website, FAQ section). With this
configuration, MINE.jar used 5.1 GB and 7.5 GB RAM for 1500 and 2000
variables respectively, then stopped at about 2.7 $\times 10^6$
comparisons without completing the task.
\par
\begin{figure*}[!ht]
\centerline{\includegraphics[width=0.75\columnwidth]{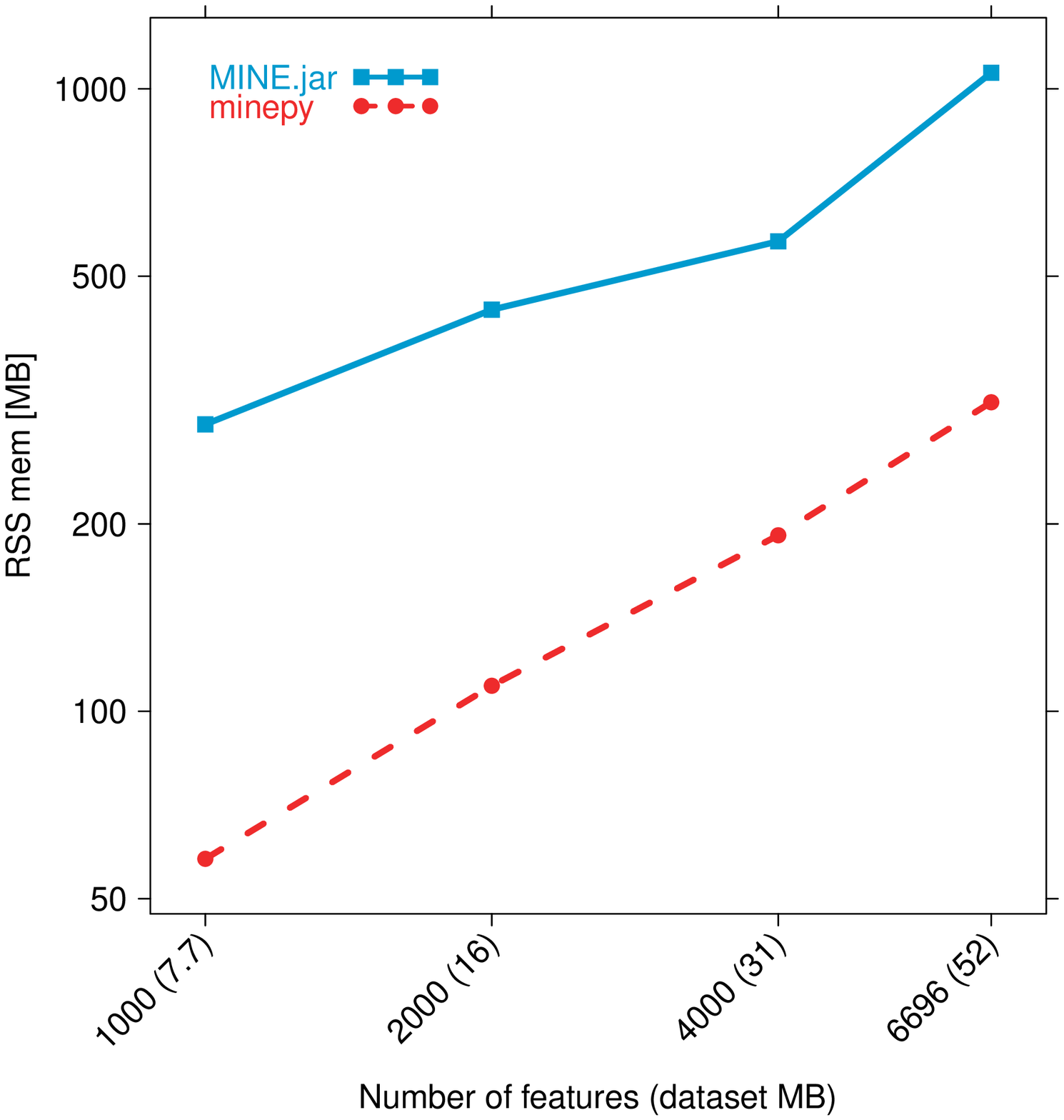}}
\caption{Comparison of memory usage of MINE.jar and minepy computing
  the four MINE statistics, the Pearson correlation coefficient and
  the non linearity index for variable 1  versus all of the Microbiome
  dataset. In parentheses, the dataset ASCII file size in megabytes
  (MB).}
  \label{fig:microbiome-mem-test}
\end{figure*}

 In comparison, memory requirement for our MINE implementations
grows much slower on the Spellman dataset. Indeed, the C++ interface has the
lowest memory footprint (less than 2 MB for 4382 variables).
%on this small size dataset, memory is mostly allocated to the Python
%environment with NumPy module loaded (about 12 MB). 
For the
Python, R, MATLAB and Octave interfaces, memory is mostly used by the
environments.
% on the Spellman dataset.
On the much more numerous
Microbiome dataset (675 samples), memory usage of minepy is
consistently smaller than MINE.jar for a MINE analysis on variable 1
vs all variables, as shown in Tab.~\ref{tab:microbiome-mem-test} and
Fig.~\ref{fig:microbiome-mem-test}.

\begin{table*}[ht]
\begin{center}
\begin{tabular}{rrrrrrrr}
  \hline
Num & Dataset~~ & MINE.jar & minepy & minerva & matlab & octave & C++ \\ 
Feat &  (kB) &  &  &  &  & &  \\ 
  \hline
  % 200 & 28 & 470,980 & 13,368 & 33,152 & 78,592 & 46,500 \\ 
  % 500 & 72 & 1,315,072 & 13,932 & 33,464 & 78,660 & 46,556 \\ 
  % 1,000 & 144 & 2,243,284 & 14,656 & 31,944 & 78,748 & 46,644 \\ 
  % 1,500 & 208 & 4.9GB & 15,516 & 32,100 & 79,096 & 46,732 \\ 
  % 2,000 & 276 & 6.9GB & 16,368 & 33,964 & 79,332 & 46,824 \\ 
  % 3,000 & 412 & -- & 17,448 & 33,264 & 80,000 & 47,004 \\ 
  % 4,382 & 600 & -- & 19,560 & 33,160 & 79,972 & 47,252 \\ 
     200 & 28 & 480.62 & 13.05 & 32.38 & 76.75 & 45.41 &  1.14\\
     500 & 72 & 2,242.00 & 13.61 & 32.68 & 76.82 & 45.46 & 1.21 \\
   1,000 & 144 & 3,798.57 & 14.31 & 31.20 & 76.90 & 45.55 & 1.30 \\
   1,500 & 208 & 5,218.90 & 15.15 & 31.35 & 77.24 & 45.64 & 1.39 \\
   2,000 & 276 & 7,670.30 & 15.98 & 33.17 & 77.47 & 45.73 &  1.49\\
   3,000 & 412 & --~~~ & 17.04 & 32.48 & 78.12 & 45.90 & 1.67 \\
   4,382 & 600 & --~~~ & 19.10 & 32.38 & 78.10 & 46.14 & 1.94 \\
   \hline
\end{tabular}
\end{center}
\caption{Memory usage, evaluated as resident set size (MB), i.e. the non-swapped physical memory used by the task, for MINE.jar, minepy, minerva, MATLAB, Octave, C++. }
\label{tab:mem}
\end{table*}
\begin{table*}[ht]
\begin{center}
\begin{tabular}{rrrrrrrr}
  \hline
Num & Dataset~~ & MINE.jar & minepy & minerva & matlab & octave & C++ \\ 
Feat & (kB) &  &  &  &  &  \\ 
\hline
% 200 & 28 & 4 & 13 & 49 & 75 & 30 \\ 
%   500 & 72 & 24 & 82 & 311 & 461 & 180 \\ 
%   1,000 & 144 & 95 & 320 & 1,224 & 1,976 & 718 \\ 
%   1,500 & 208 &  330 & 718 & 2,825 & 4,161 & 1,621 \\ 
%   2,000 & 276 &  671 & 1,275 & 4,936 & 7,617 & 2,869 \\ 
%   3,000 & 412 &  -- & 2,885 & 11,185 & 16,689 & 6,467 \\ 
%   4,382 & 600 &  -- & 6,115 & 24,307 & 35,810 & 14,147 \\
200 & 28 & 6 & 13 & 49 & 75 & 30 & 11\\
   500 & 72 & 37 & 82 & 311 & 461 & 180 & 68\\
   1,000 & 144 & 146 & 320 & 1,224 & 1,976 & 718 &  280\\
   1,500 & 208 & 349 & 718 & 2,825 & 4,161 & 1,621 & 653\\
   2,000 & 276 & 664 & 1,275 & 4,936 & 7,617 & 2,869 & 1,119\\
   3,000 & 412 &  --~~~ & 2,885 & 11,185 & 16,689 & 6,467 & 2,457\\
   4,382 & 600 &  --~~~ & 6,115 & 24,307 & 35,810 & 14,147 & 5,199\\
   \hline
\end{tabular}
\end{center}
\caption{Computing time comparisons (secs) for MINE.jar, minepy, minerva,  MATLAB, Octave and C++ interfaces, 
on the Spellman dataset (all pairs); as elapsed real (wall clock) time used by the process.}
\label{tab:real.times}
\end{table*}
\begin{table*}[ht]
\begin{center}
\begin{tabular}{rrrr}
  \hline
Num  & Dataset~~ & MINE.jar & minepy \\ 
 Feat & Dim (MB) &  &  \\ 
  \hline
1,000 & 8 & 295,984 & 59,332 \\ 
  2,000 & 16 & 452,356 & 112,536 \\ 
  4,000 & 31 & 582,480 & 196,352 \\ 
  6,696 & 52 & 1,086,504 & 321,184 \\ 
   \hline
\end{tabular}
\end{center}
\caption{Comparison of memory usage (kB) for MINE.jar and minepy 
  computing the four MINE statistics,
  the Pearson correlation coefficient and the non linearity index for variable
  1 versus all of the Microbiome dataset. Memory is evaluated as 
  resident set size (RSS), i.e. the non-swapped physical memory used by the 
  task,  for increasing number of features.}
  \label{tab:microbiome-mem-test}
\end{table*}

We also evaluated on the same data the effect of enabling
parallelization in minerva. This can be managed natively in R by using
functionalities of the \textit{parallel} package \cite{rcore12R}.
An example of how to enable multicore computation is: 

%\begin{minipage}{\textwidth}
\tiny
\begin{verbatim}
> # load the minerva package
> library(minerva)
> # load the Spellman dataset (available in minerva)
> data(Spellman)
> # run MINE on variable ``time'' vs all, with 8 cores 
> res <- mine(x=Spellman, master=1, alpha=0.67, C=15, n.cores=8)
\end{verbatim}
\normalsize
%\end{minipage}

\begin{figure*}[!ht]
\centerline{\includegraphics[width=0.75\columnwidth]{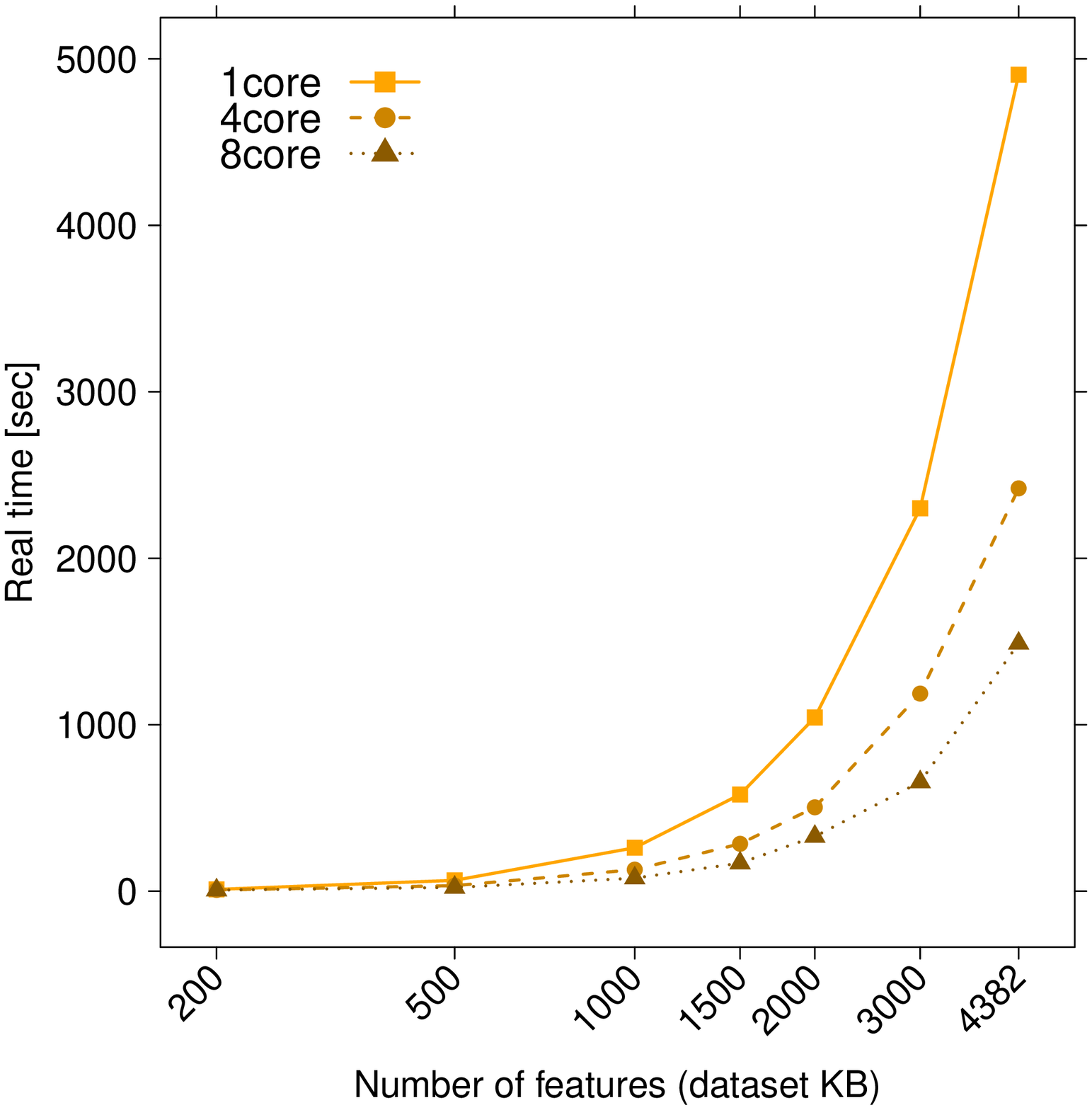}}
\caption{Elapsed time used by the process (in seconds) versus
  increasing number of features (log scale) to compute the MINE
  statistics for all pairs of features of the CDC15 Spellman yeast
  dataset. }
  \label{fig:rparallel}
\end{figure*}

In the experiment we computed the MINE statistics with 1, 4, or 8
cores for all pairs of features of the Spellman dataset, obtaining the
plot in Fig.~\ref{fig:rparallel}. For 4382 features, computing time is
reduced from about 5000 secs with one core to 1500 with 8 cores. 

\newpage
%\mbox{}
%\newpage
\noindent
\paragraph{2B.} To evaluate scalability, an experiment of MPI
parallelization of minerva was performed with the Rmpi package on the
high-performance Linux computing cluster FBK-KORE. Up to 100 cores
(described in Subsection \ref{ssec:kore}) were employed in the tests,
managed by the Sun Grid Engine (SGE) batch-queuing system. The MINE
analysis (four MINE statistics) was applied to 5 instances of the
Spellman dataset, in ``all vs all'' mode and 10 different configurations
with increasing number of cores up to 100. Speedup was defined
as: $\textrm{Speedup}(p)= T_{1}/T_{p}$, where $T_{p}$ is the time cost
of the algorithm on $p$ processes.  As in 2A, computing time was
considered as wall clock time.
\par
\begin{figure*}[!ht]
  \centerline{\includegraphics[width=0.8\columnwidth]{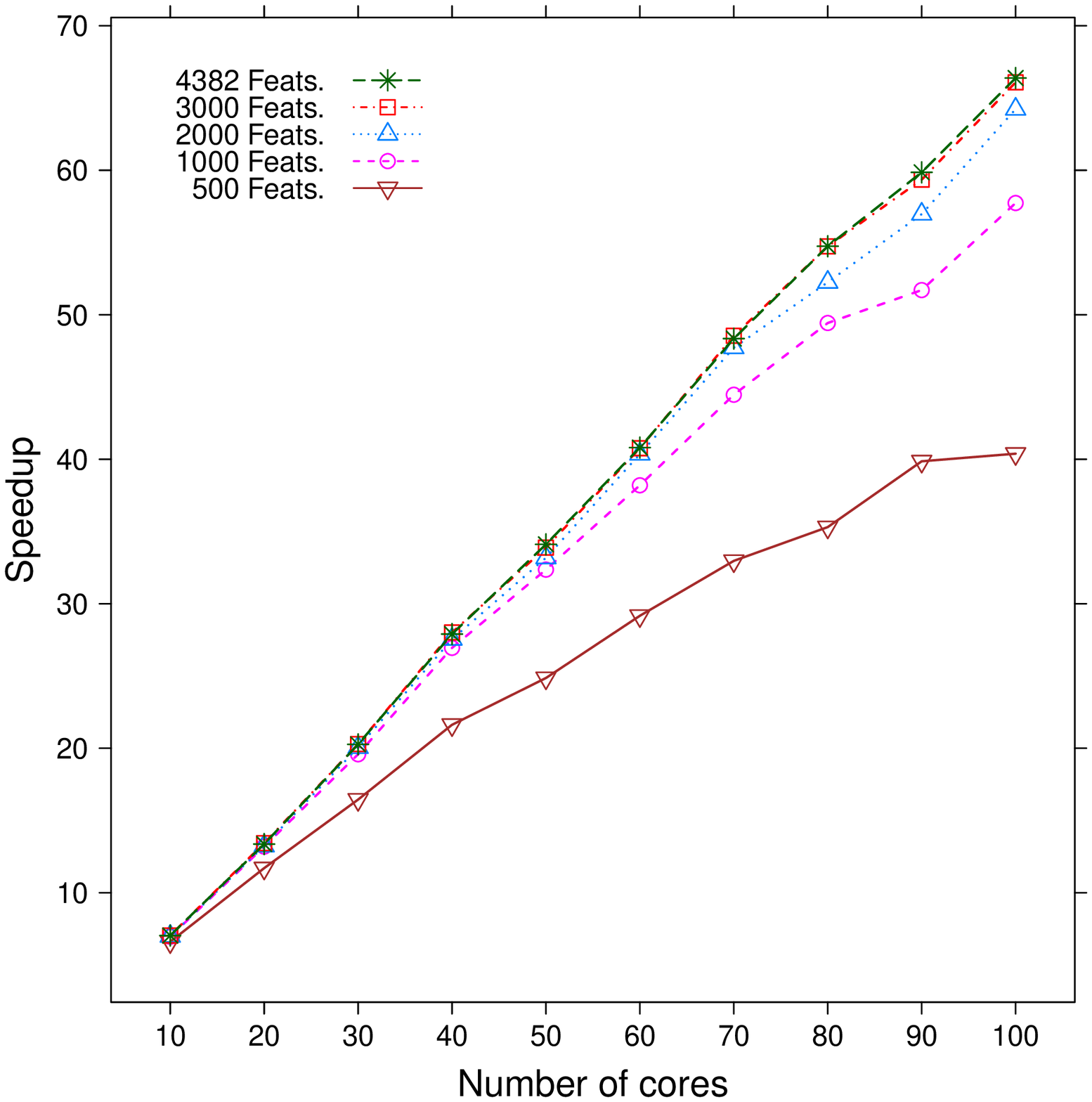}}
  \caption{Speedup plot for increasing number of cores from 10 to
    100. The MINE statistics were computed for all pairs and
    increasing feature set sizes on the CDC15 Spellman yeast dataset
    ($\alpha$=0.6). }
  \label{fig:spellman-speedup}
\end{figure*}

The speedup is higher for larger datasets, with the curve for 4382
features almost overlapping that for 3000 (saturation effect).  A
speedup of about 70 was achieved with 100 cores for 4382
features. Results are displayed in Fig.~\ref{fig:spellman-speedup}.

\vspace{1cm}

\noindent
\paragraph{2C.} The scaling properties of MINE as a function of the
sample size $n$ and the $\alpha$ parameter are important in practical
applications. As introduced in \cite{reshef11detecting}, $\alpha$
defines the maximum grid size $B$ according to the relation
$B(n)=n^\alpha$. Finer grids correspond obviously to higher
computational costs. We empirically evaluated how computing time grows
for varying $\alpha$ and $n$ by considering a MINE analysis for two
variables on datasets of increasing size, here simulated by two
uniformly distributed random vectors of increasing length. The R
interface minerva was applied on a standard laptop system (sw and hw
details in Sec.~\ref{ssec:laptop}). 

\begin{figure*}[!ht]
  \centerline{\includegraphics[width=0.75\columnwidth]{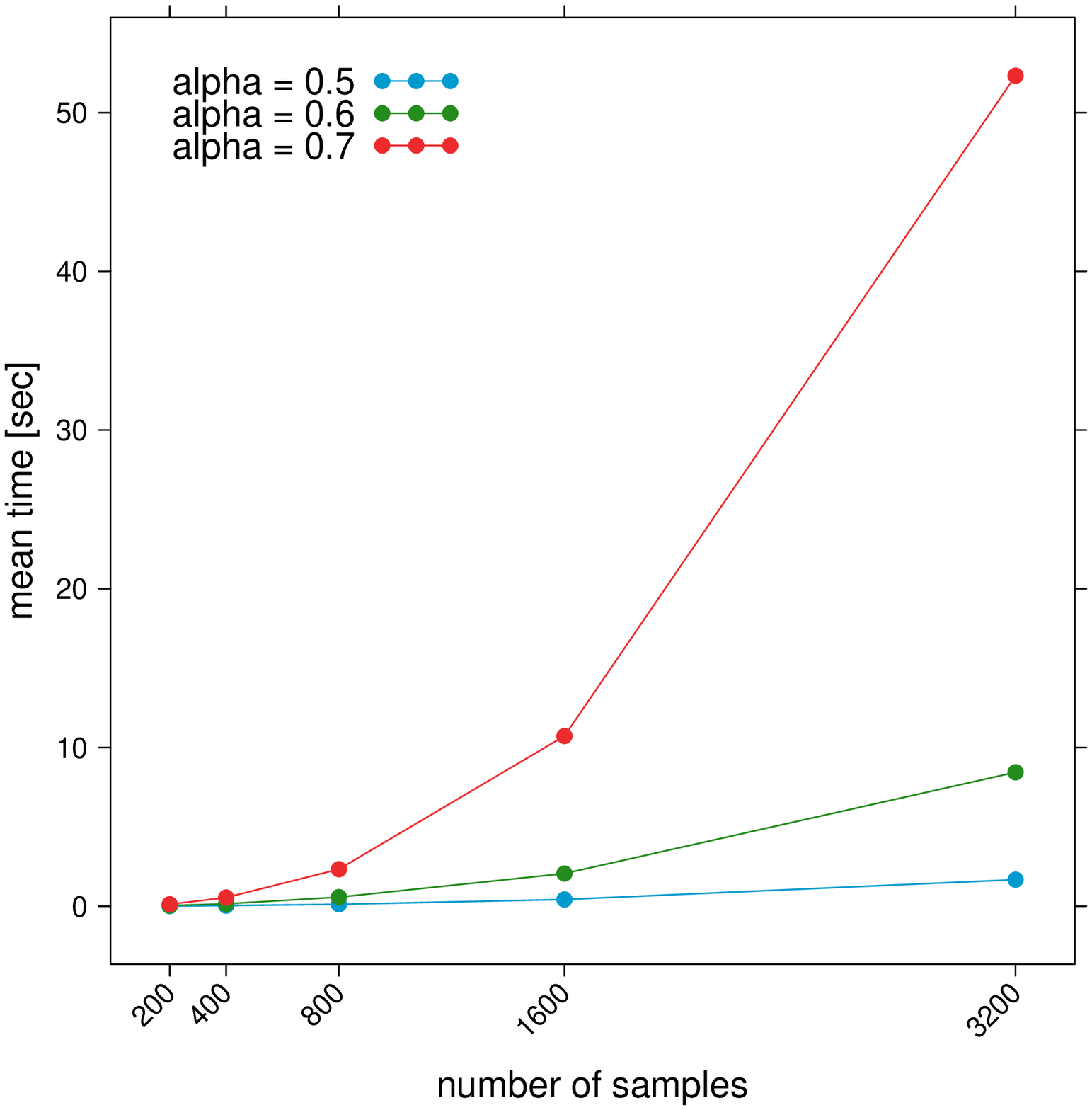}}
  \caption{Average of elapsed time on 100 repetitions of MINE analysis
    with the minerva package on two uniformly distributed random
    variables for an increasing number of samples and $\alpha=0.5,
    0.6, \textrm{and } 0.7$.}
  \label{fig:sample-size}
\end{figure*}

Fig.~\ref{fig:sample-size}
displays the mean elapsed time for 100 replicates for 5 different
sample sizes. The results confirm how critical is the choice of
$\alpha$ on computing time: the standard $\alpha=0.6$ proposed in
\cite{reshef11detecting} requires less than 15\% of the time needed
for $\alpha=0.7$.

Due to the linearity in computing MINE statistics on
$p$ pairs of variables, Fig.~\ref{fig:sample-size} can be used to
derive a rough estimate of the total time required to perform a MINE
computation on a given dataset.

\vspace{1.5cm}

\section{Transcriptomic datasets}

In this section, we describe the high-throughput transcriptomic
datasets used for the additional experiments reported in Tab.~1 of the
main paper. In both experiments, we set $c=15$ and $\alpha=0.6$;
hardware and software configurations are specified in Subsection
\ref{ssec:mlbio}.

\vskip .1in
\noindent
\paragraph{3A} \textbf{Human brain transcriptome dataset (GSE25219).}
This exon array dataset \cite{kang11spatio} was obtained from 1340
samples, collected from multiple regions of 57 postmortem human
brains, spanning from embryonic development to late adulthood, 31
males and 26 females of different ethnicities. The samples were
spotted on a Affymetrix Human Exon 1.0 ST Array, and core and unique
probe sets were summarized, representing 17,565 mainly protein-coding
genes, into gene-level information (transcript gene version). Note
that at the same accession number GSE25219 an exon version of the data
is also provided, but we used the gene version for Tab.~1 of the main
paper.  For replicability, the dataset formatted for the MINE analysis
is available at
\url{http://sourceforge.net/projects/minepy/files/data/GSE25219.csv}.

\vskip .1in
\noindent
\paragraph{3B} \textbf{Non-small cell lung cancer RNA-seq dataset
  (GSE34914).}  The data were generated by an RNA-sequencing study of
non-small cell lung cancer (NSCLC) \cite{kalari12deep}. The cohort
includes 8 lung adenocarcinomas with mutant KRAS in lung tumors and 8
lung adenocarcarcinomas without KRAS mutation. One tumor sample
without KRAS mutation was run twice for QC evaluation.  The samples
were sequenced on an Illumina Genome Analyzer II. The raw read counts
from BWA alignment are available for 22,316 genes, but we filtered out
1,894 features having variance smaller than $10^{-5}$. For
replicability, the dataset formatted for the MINE analysis is
available at
\url{http://sourceforge.net/projects/minepy/files/data/GSE34914.csv}.

\section{Hardware/Software configurations}
\label{sec:system}

\subsection{Experiments: 1A, 2A}
\begin{itemize}
\item 8 cores Intel\textsuperscript{\textregistered} Xeon\textsuperscript{\textregistered} E5540 2.53 GHz 64 bit workstation
  (``krk''), with 12 GB RAM
\item Linux 2.6.18
\item Red Hat 4.1.2
\item GCC 4.1.2  
\item Java\textsuperscript{\texttrademark} 1.7.0 
\item Python 2.7.3 and NumPy 1.6.2
\item R 2.15.1
\item MATLAB\textsuperscript{\textregistered} 7.6.0.324 (R2008a)
\item GNU Octave 3.0.5
\item MINE.jar 1.0.1
\item minepy 0.3.5
\item minerva 1.1
\end{itemize}

\subsection{Experiments: 1B, 2B}
\label{ssec:kore}
These experiments were run on a high-performance computing Linux
cluster (``FBK-KORE''), with more than 700 cores and 200 TB disk
space. A queue was reserved for the experiments, in which each
computing node was equipped with:

\begin{itemize}
\item quad-core Intel\textsuperscript{\textregistered} Xeon\textsuperscript{\textregistered} E5420 2.5 GHz, reserving 2 GB RAM
\item Red Hat 4.4.6-4
\item GCC  4.4.6
\item Open MPI 1.5.4
\item Rmpi 0.6-1
\item Python 2.7.3 and NumPy 1.6.1
\item R 2.15.1
\item minepy 0.3.5
\item minerva 1.1
\end{itemize}

\subsection{Experiment: 2C}
\label{ssec:laptop}
\begin{itemize}
\item Intel\textsuperscript{\textregistered} Core\textsuperscript{\texttrademark} 2 Duo  3.06 GHz laptop (``Dna'') with 4GB RAM
\item Mac OS X Lion 10.7.4
\item GCC 4.2.1 (Based on Apple Inc. build 5658) (LLVM build 2336.9.00)
\item R 2.15.1
\item minerva 1.1
\end{itemize}

\subsection{Experiments: 3A, 3B}
\label{ssec:mlbio}
 \begin{itemize}
 \item 24 core Intel\textsuperscript{\textregistered} Xeon\textsuperscript{\textregistered} E5649 CPU 2.53GHz workstation (``mlbio'') with 47 GB RAM
 \item Linux 2.6.32
 \item Red Hat 4.4.6
 \item GCC 4.4.6
 \item Java 1.7.0
 \item Python 2.6.6 and Numpy 1.6.0
 \item R 2.15.1
 \item MINE.jar 1.0.1
\item minepy 0.3.5
\item minerva 1.1
 \end{itemize}

\section{Examples}

We provide in this section examples of usage of interfaces based on
the libmine C library. All datasets considered in this section are
available at \url{http://sourceforge.net/projects/minepy/files/data},
and at the MINE website \url{http://www.exploredata.net}. All examples 
are based on minepy 0.3.5 and minerva 1.1.

\subsection{Spellman dataset}
\label{ssec:spellman}

In this example we will first compute the MINE statistics on the
Spellman dataset, analyzing the association between variable ``time''
and each of the other variables, printing the MIC value of ``time'' vs
``YAL001C''. We use the same parameter configuration as in
\cite{reshef11detecting}: $\alpha=0.67$ and $c=15$. The dataset is
available at\newline
\url{http://sourceforge.net/projects/minepy/files/data/Spellman.csv}
or within the minerva package for R users.

\subsubsection{Python API (included in minepy)} 
Download and install the latest minepy module from the project page
(\url{http://minepy.sourceforge.net}).
\tiny
\begin{verbatim}
>>> # import the numpy module
>>> import numpy as np
>>> # import the minepy module
>>> from minepy import MINE
>>> # load the Spellman dataset
>>> spellman = np.genfromtxt('Spellman.csv', delimiter=',')[:, 1:]
>>> # initialize results values with an empty list
>>> res = []
>>> for i in range(1, len(spellman)):
...     # build the mine object
...     mine = MINE(alpha=0.67, c=15)
...     # compute the MINE statistics
...     mine.score(spellman[0], spellman[i])
...     res.append(mine)
... 
>>> # print MIC of 'time' vs 'YAL001C'
>>> res[0].mic()
0.6321377231641834
\end{verbatim}
\normalsize

\subsubsection{\texttt{mine} Application (included in minepy)}
Download and install the latest mine application from the minepy homepage (\url{http://minepy.sourceforge.net}).
\tiny
\begin{verbatim}
$ mine Spellman.csv -a 0.67 -c 15 -m 1 -o Spellman_MINE.txt
\end{verbatim}
\normalsize
The MINE statistics will be stored in the \texttt{Spellman\_MINE.txt} file.

\subsubsection{R API (minerva)}
Download and install the latest minerva package from CRAN \\
(\url{http://cran.r-project.org/web/packages/minerva/index.html}).%
\tiny
\begin{verbatim}
> # load the minerva package
> library(minerva)
> # load the Spellman dataset 
> data(Spellman)
> # compute the MINE statistics 
> res <- mine(x=Spellman, master=1, alpha=0.67, C=15)
> # print 'time' vs 'YAL001C'
> res[["MIC"]]["YAL001C",]
[1] 0.6321377
\end{verbatim}
\normalsize%
\subsubsection{MATLAB/Octave API (included in minepy library)}
Download and install the MATLAB interface from  the latest minepy package (\url{http://minepy.sourceforge.net}).
\tiny
\begin{verbatim}
>> % load the Spellman dataset
>> spellman = csvread('Spellman.csv', 0, 1);
>> n = length(spellman);
>> % preallocate and initialize to zero the results array
>> res(1:n-1) = struct('mic',0,'mas',0,'mev',0,'mcn',0);
>> % compute the MINE statistics
>> for i = 2:n
res(i-1) = mine(spellman(1,:),spellman(i,:), 0.67, 15);
end
>> % print 'time' vs 'YAL001C' 
>> res(1)

ans = 

    mic: 0.6321
    mas: 0.2537
    mev: 0.6321
    mcn: 3
\end{verbatim}
\normalsize

\subsection{Microbiome dataset}
In this example we will compute the MINE statistics (``all vs all''
study) on the Microbiome dataset by using the mine Application. We use
the parameter configuration $\alpha=0.551$ and $c=10$. The dataset is available at\\
\url{http://sourceforge.net/projects/minepy/files/data/Microbiome.csv}.

Note that in this experiment the computational time may be very high
on a standard workstation. First download and install the latest
minepy package from \url{http://minepy.sourceforge.net}.  To launch
the analysis:
\tiny
\begin{verbatim}
$ mine Microbiome.csv -a 0.551 -c 10 -o Microbiome_MINE.txt
\end{verbatim}
\normalsize
To compute the MINE statistics between variable \#100 versus all
the others we can use the ``master variable'' option (-m):
\tiny
\begin{verbatim}
$ mine Microbiome.csv -a 0.551 -c 10 -m 100 -o Microbiome_MINE.txt
\end{verbatim}
\normalsize
The MINE statistics between variables \#100 and \#200 can be also
directly computed by using the ``pair'' option (-p):
\tiny
\begin{verbatim}
$ mine Microbiome.csv -a 0.551 -c 10 -p 100 200 -o Microbiome_MINE.txt
\end{verbatim}
\normalsize
Output statistics for the examples above will be stored in the
\texttt{Microbiome\_MINE.txt} file.

\subsection{Baseball dataset}
In this example we will compute the MINE statistics (``all vs all''
mode) on the Baseball dataset with the mine Application with
$\alpha=0.7$ and $c=15$. The dataset is available at\\
\url{http://sourceforge.net/projects/minepy/files/data/MLB2008.csv}. Note
that in this experiment the computational time may be very high on a
standard workstation. First download and install the latest minepy package from
\url{http://minepy.sourceforge.net}.  To launch the analysis:
\tiny
\begin{verbatim}
$ mine MLB2008.csv -a 0.7 -c 15 -o MLB2008_MINE.txt
\end{verbatim}
\normalsize
The statistics will be stored in the \texttt{MLB2008\_MINE.txt} file.

\newpage

\subsection{C++ example} 
In this example we will compute the MINE statistics between $x = \{0, 0.001, \dots, 1\}$
and $y=\sin(10 \pi x) + x$. 
Download and untar the latest minepy module from the project page
(\url{http://minepy.sourceforge.net}). This example is located in  minepy-X.Y.Z/examples/cpp\_example.cpp (where X.Y.Z  is the current version of minepy).
\tiny
\begin{verbatim}
#include <cstdlib>
#include <cmath>
#include <iostream>
#include "cppmine.h"

using namespace std;

int
main (int argc, char **argv)
{
  double PI;
  int i, n;
  double *x, *y;
  MINE *mine;
  
  PI = 3.14159265;
  
  /* build the MINE object with exceptions management */
  try {    
    mine = new MINE(0.6, 15);
  }
  catch (char *s) {
      cout << "WARNING: " << s << "\n";	
      cout << "MINE will be set with the default parameters," << "\n";
      cout << "alpha=0.6 and c=15" << "\n";
      mine = new MINE(0.6, 15);
  }
  
  /* build the problem */
  n = 1001;
  x = new double [n];
  y = new double [n];
  for (i=0; i<n; i++)
    {
      /* build x = [0, 0.001, ..., 1] */
      x[i] = (double) i / (double) (n-1);

      /* build y = sin(10 * pi * x) + x */
      y[i] = sin(10 * PI * x[i]) + x[i]; 
    }
  
  /* compute score */
  mine->compute_score(x, y, n);
  
  /* print mine statistics */
  cout << "MIC: " << mine->get_mic() << "\n";
  cout << "MAS: " << mine->get_mas() << "\n";
  cout << "MEV: " << mine->get_mev() << "\n";
  cout << "MCN: " << mine->get_mcn() << "\n";

  /* delete the mine object */
  delete mine;
  
  /* free the problem */
  delete [] x;
  delete [] y;
  
  return 0;
}
\end{verbatim}
\normalsize
On a standard Linux machine, open a terminal, go into the example folder  (examples/) and compile the code:
\tiny
\begin{verbatim}
$ g++ -O3 -Wall -Wno-write-strings cpp_example.cpp \
../minepy/libmine/cppmine.cpp ../minepy/libmine/core.c \
 ../minepy/libmine/mine.c -I../minepy/libmine/
\end{verbatim}
\normalsize
Run the example by typing:
\tiny
\begin{verbatim}
$ ./a.out
MIC: 0.999999
MAS: 0.728144
MEV: 0.999999
MCN: 4.584963
\end{verbatim}
\normalsize
\subsection{Other examples}

Short examples are available in the on line documentation about the
APIs and the mine application at:
\url{http://minepy.sourceforge.net/docs}. For example, for v0.3.5, the
examples can be found at:
\tiny
\begin{itemize}
  \item[    Python]:
    \url{http://minepy.sourceforge.net/docs/0.3.5/python.html}
 \item[    C]: \url{http://minepy.sourceforge.net/docs/0.3.5/c.html}
 \item[    C++]: \url{http://minepy.sourceforge.net/docs/0.3.5/cpp.html}
 \item[    MATLAB/Octave]:
    \url{http://minepy.sourceforge.net/docs/0.3.5/matlab.html}
 \item[    mine application]:
    \url{http://minepy.sourceforge.net/docs/0.3.5/application.html}
  \end{itemize}

  For minerva, the most updated documentation and examples are available at
  \begin{itemize}
   \item [R]: \url{http://cran.r-project.org/web/packages/minerva/minerva.pdf}
  \end{itemize}
\end{document}